\documentclass[conference]{IEEEtran}

\usepackage{amsmath}
\usepackage[numbers]{natbib}
\usepackage{microtype}
\usepackage{hyperref}
\usepackage{url}
\usepackage{booktabs}
\usepackage{graphicx}
\usepackage[table,xcdraw]{xcolor}
\usepackage{textgreek}
\usepackage{tipa}
\usepackage{tikz}
\usepackage{paralist}
\usepackage[shortlabels]{enumitem}
\usepackage{epigraph}
\usepackage[most]{tcolorbox}
\usepackage{tabularx}
\usepackage{soul}
\usepackage{subcaption}
\usepackage{longtable}
\usepackage{cleveref}
\usepackage{placeins}

\newtcolorbox[list inside=prompt,auto counter,number within=section]{prompt}[1][]{
    colbacktitle=black!60,
    fonttitle=\small,
    coltitle=white,
    fontupper=\footnotesize,
    boxsep=4pt,
    left=0pt,
    right=0pt,
    top=0pt,
    bottom=0pt,
    boxrule=1pt,
    #1,
}

\newcommand{\li}[1]{\textit{latent index #1}}
\newcommand{\Li}[1]{\textit{Latent index #1}}
\DeclareRobustCommand{\appref}[1]{\begingroup\renewcommand{\sectionautorefname}{Appendix}\autoref{#1}\endgroup}

\makeatletter
\patchcmd{\epigraph}{\@epitext{#1}}{\itshape\@epitext{#1}}{}{}
\makeatother

\definecolor{darkblue}{rgb}{0, 0, 0.5}
\hypersetup{colorlinks=true, citecolor=darkblue, linkcolor=darkblue, urlcolor=darkblue}

\begin{document}

\title{On the Interpretability of Whisper Encodings Using Sparse Autoencoders}

\author{
\IEEEauthorblockN{Dan Pluth}
\IEEEauthorblockA{Vail Systems, Inc.}
\and
\IEEEauthorblockN{Zachary Nicholas Houghton}
\IEEEauthorblockA{Vail Systems, Inc. \\ University of Oregon}
\and
\IEEEauthorblockN{Yu Zhou}
\IEEEauthorblockA{Vail Systems, Inc.}
\and
\IEEEauthorblockN{Vijay K. Gurbani}
\IEEEauthorblockA{Vail Systems, Inc.}
}

\maketitle

\begin{abstract}
While deep transformer-based models have advanced rapidly, their internal mechanisms remain largely a mystery. Recent work has prioritized understanding text-based transformer models, leaving ASR systems largely unexplored. In order to address this gap, we examine the internal representations of Whisper's encoder using a sparse autoencoder. We find diverse monosemantic features across linguistic and non-linguistic boundaries, spanning a hierarchy from phonetic to semantic representations, and conduct a causal feature-steering campaign across this hierarchy, including cross-lingual steering. We further find that steering is more reliable for higher-level features than lower-level ones, an asymmetry that may reflect redundant encoding of lower-level information. Altogether, this work demonstrates that Whisper's encoder represents a surprisingly rich hierarchy of linguistic information that extends well beyond what is strictly necessary for transcription.
\end{abstract}

\section{Introduction} \label{sec:intro}

In the last few years, advances in the Transformer architecture have led to
models with the ability to produce text, audio, image, and video. These
impressive capabilities come with an increased need to understand their internal
representations. While there have been many studies focused on understanding the internal representations of LLMs \citep[e.g.,][]{houghton2025role,houghton2026verbup}, fewer studies have examined the internal representations of audio-based models, such as automatic speech recognizers. 

One such ASR model is Whisper, which is capable of transcribing and translating spoken language in a wide variety of languages \citep{radford2023robust}. Whisper is comprised of an encoder, which maps raw audio into a dense representation, and a decoder, which autoregressively generates the transcript from that representation. This division of labor means the decoder, not the encoder, is responsible for the actual language modeling. The encoder's training objective imposes no obvious requirement to represent anything beyond the acoustic and phonetic information that the decoder needs to do its job: a purely phonetic code would, in principle, already suffice. Yet there is already some evidence that the encoder represents more than this minimum: background noise information, for instance, appears to condition the decoder's output even though it is not itself part of the transcript \citep{gong23d_interspeech}. This raises the possibility that the encoder more broadly represents linguistic information that is not strictly necessary for transcription, but that nonetheless facilitates it. The present study asks whether the Whisper encoder represents anything beyond surface-level acoustic information.

One form of mechanistic interpretability well suited to this kind of question is the sparse autoencoder (SAE), which decomposes a model's dense internal representations into a sparse, more human-interpretable set of latent features \citep{bricken2023monosemanticity}. SAEs have been successful in identifying monosemantic, interpretable features within the densely encoded representations produced by intermediate layers of large language models, finding structure in previously uninterpretable machine output. To date, however, this technique has been applied primarily to large language models operating on text, with only limited application to audio-based models \citep{aparin2026audiosae}.

This raises two open questions. First, it remains unclear how well SAE-based interpretability, developed and validated almost entirely on text-based models, generalizes to an audio-based encoder like Whisper's. Second, and independent of how well the method itself generalizes, it remains unclear whether Whisper's encoder represents any high-level linguistic categories at all, given that its training objective does not obviously require it to do so.

If SAE-based interpretability does not generalize well to an audio encoder, decomposing Whisper's representations should yield latents that are largely uninterpretable, with no consistent correspondence to identifiable properties of the audio or its transcript; if it generalizes well, we should instead recover latents that reliably track specific, nameable properties of the input, just as has been found for text-based models. Separately, if the encoder represents only what transcription strictly requires, any such latents should cluster almost entirely around low-level phonetic and acoustic distinctions, with any apparent higher-level structure being weak, sparse, or explainable as an artifact of correlated acoustic cues. 

In this work, we apply a SAE to Whisper's encoder to address both questions at once. The key contributions of this work are threefold:

\begin{enumerate}
    \item We show that Whisper's encoder represents a hierarchy of linguistic information (phonetic, lexical, syntactic, morphological, and semantic) that goes well beyond what the encoder needs to extract for transcription alone.
    \item We conduct a causal feature-steering campaign spanning the full hierarchy, characterizing where steering succeeds and where it fails.
    \item Our findings speak to how well SAE-based interpretability methods generalize to Whisper's encoder, an open question given the limited prior application of the method to audio-based models.
\end{enumerate}

\section{Related Work}
\label{sec:related-work}

There has been a substantial body of research attempting
to understand the internal representations of large language models \citep{tenney2019bert, haley2020bert,
li2021neural, lasri2022subject, yao2025both, misra2024language,
houghton2025multi, pan2025explicit, houghton2025role}. These results typically rely on examining the probability distribution of output tokens in carefully designed studies. For example, \citet{houghton2025role} examined whether ordering preferences of certain linguistic constructions in LLMs, such as binomials, are driven by abstract knowledge or whether they are instead driven by item-specific memorization.

Recently, however, researchers have found success with mechanistic
interpretation methods for inspecting the internal workings of models. One such
approach is the use of sparse autoencoders (SAEs) to decompose the densely
encoded information of a model into sparse monosemantic features. Much of this
work has focused on interpretability in text-based Large Language Models (LLMs) as a means
to better understand what information is being encoded in them \citep{bricken2023monosemanticity,huben2024sparse, templeton2024scaling,
gao2025scaling}. For example,
\citet{bricken2023monosemanticity} demonstrated the effectiveness of training a
sparse autoencoder on the internal activations of a 1-layer transformer model.
They found that a sparse autoencoder trained this way learns monosemantic
features such as text written in Arabic. Further, \citet{huben2024sparse} built
upon this and demonstrated the effectiveness of sparse autoencoders on even more
complex language models. Specifically, they demonstrated that a sparse
autoencoder trained on the internal activations of Pythia
\citep{biderman2023pythia} learns human-interpretable features, such as a
feature for court cases and a feature for last names.
\citet{sun2025} found that a SAE is capable of learning intrinsically dense features of an LLM, while
\citet{chanin2025a} examined SAE feature splitting and absorptions.

While these studies demonstrate that SAEs are a useful tool for interpreting the internal representations of text-based LLMs, it remains less clear whether this approach generalizes to other modalities, such as end-to-end ASR models that take audio rather than text as their input, and in particular whether it can identify higher-level linguistic structure rather than lower-level acoustic and non-linguistic content. Some prior work has probed the internal representations of Whisper by building a classifier from each layer of the encoder \citep{gong23d_interspeech}, and others have applied a sparse autoencoder to a speaker recognition model \citep{11005361}.

Separately, layer-wise probing of wav2vec2, a self-supervised speech model that, unlike Whisper, has no separate decoder and is instead fine-tuned for ASR via a linear CTC output head rather than autoregressive decoding, has shown that acoustic and phonetic information is organized hierarchically across the depth of the network, with earlier layers capturing low-level acoustic properties and later layers capturing increasingly abstract information up to word identity \citep{pasad2021layerwise}, mirroring similar depth-wise findings in text-based LLMs \citep{tenney2019bert}. Because wav2vec2 has no downstream decoder to which language modeling can be offloaded, these findings do not directly speak to whether an ASR encoder whose decoder performs the language modeling, as in Whisper, would represent linguistic structure beyond what transcription requires. Even on its own terms, however, this hierarchy does not extend to higher-level conceptual representations: wav2vec2 representations correlate only modestly with word-similarity judgments, well below dedicated semantic embedding models, a level of correlation that falls short of establishing conceptual representations beyond word identity.

Concurrent work has applied SAEs directly to Whisper and HuBERT, another self-supervised speech model similar in spirit to wav2vec2 \citep{hsu2021hubert}, finding that latents capture acoustic, non-linguistic, and speaker-related structure, including environmental sounds and speaker traits such as accent and emotion \citep{aparin2026audiosae}. That work targets low-level acoustic and non-linguistic features rather than the higher-level lexical, morphological, or semantic structure examined here. There is therefore no convincing demonstration to date that an ASR encoder's representations extend to higher-level linguistic structure. Our work addresses this gap.

\section{Method}
\label{sec:setup}

\begin{figure*}[!t] \centering
\includegraphics[width=1\textwidth]{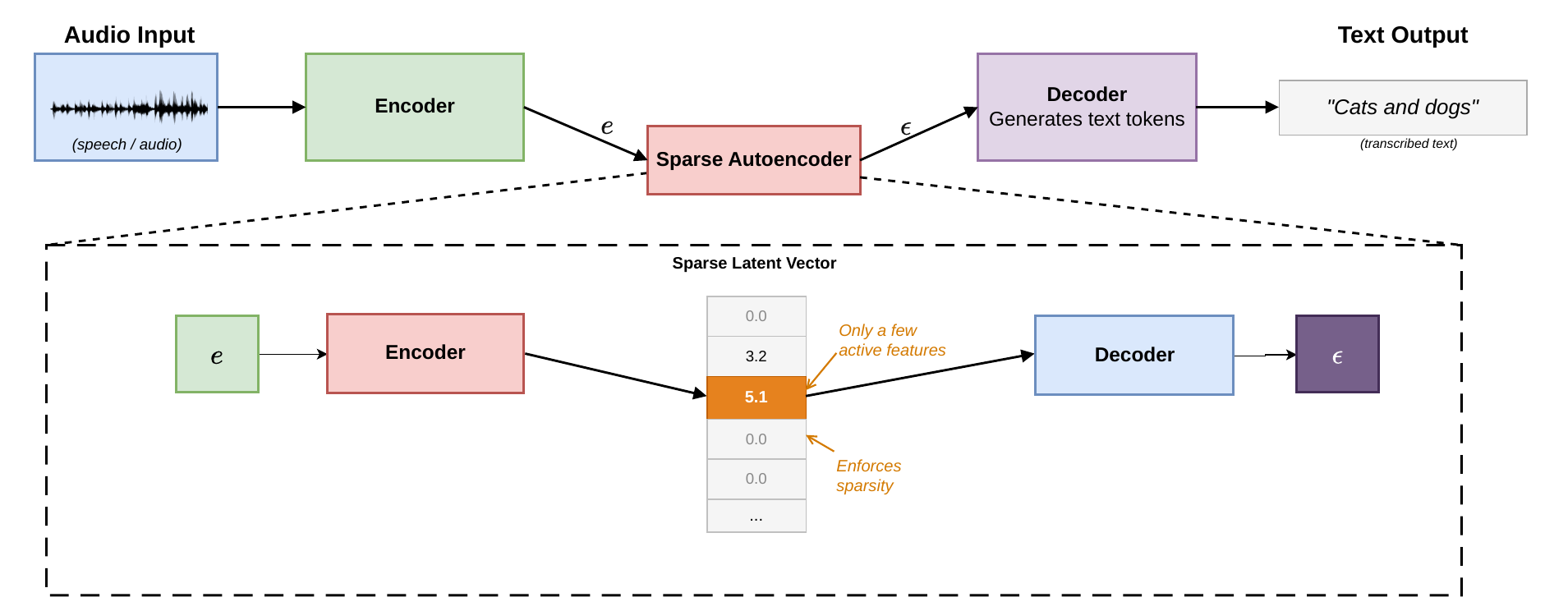}
\caption{Diagram of the Whisper architecture. The SAE is trained based on the
outputs of the Whisper encoder. When steering is performed, the reconstructed embedding
can be used as an input to the Whisper decoder to generate manipulated text.}
\label{fig:whisper_arch}
\end{figure*}

In order to examine the internal representations of Whisper, a SAE was applied
immediately after the final layer of the encoder.
  Since the final layer of the encoder is directly processed by the decoder, it potentially contains the most important information in order for the decoder to accurately transcribe the audio.\footnote{Although in
  principle, the SAE could be situated between any layers in the encoder or
  decoder.}
  \autoref{fig:whisper_arch} shows the Whisper architecture with the SAE inserted. This analysis focuses entirely on the latent space of the SAE, with the Whisper decoder only being utilized to examine the effects on the transcript after modifying the latent values.

We trained the SAE using a framework adapted from OpenAI's sparse autoencoder\footnote{\url{https://github.com/openai/sparse_autoencoder}.}, which implements a k-Sparse Autoencoder \citep{makhzani2014ksparseautoencoders, gao2025scaling}. The
SAE consists of a linear encoder with a TopK activation and a linear decoder that reconstructs the Whisper embedding space, mapping Whisper base's 512-dimensional encoding to a 16,000-dimensional hidden layer and back down to 512 dimensions. The model was trained with a TopK
activation (K=45) to constrain the number of active latents, creating a total of 16.4 million trainable parameters\footnote{2 layers with $512*16000$ parameters each.}.
The SAE is trained to minimize mean squared error (MSE) between its input and reconstructed output.\footnote{The training code, analysis code, analysis dataset and latent examples with automated labels can be found in this repo: \url{https://github.com/znhoughton/sae-whisper-rtc-2026}.}

Since Whisper takes an input of 30 seconds of audio, all audio files are
padded or trimmed to 30 seconds by default. To avoid training the SAE on padded frames, they are
identified and removed during SAE training. The final input data to the
SAE is specifically the individual frame-level Whisper embedding cut to the audio length. A given
audio will generate up to 1500 frames (30 seconds worth) of training data.

\subsection{Dataset}
\label{sec:dataset}

The data used to train the SAE comprised several different datasets, since we expected data
diversity to be critical to developing a rich set of latent features.
This dataset comprises LJSpeech \citep{ljspeech17}, LibriSpeech
\citep{librispeech}, Voxceleb 1 \citep{Nagrani17}, Mozilla Common Voice English
\citep{ardila-etal-2020-common},
SLR39 \citep{morgan2006west},
SLR67 \citep{mena_2019}, SLR61 and
SLR71-SLR75 \citep{guevara-rukoz-etal-2020-crowdsourcing},
Musan \citep{musan2015}, the Zamia Speech corpus\footnote{\url{https://github.com/gooofy/zamia-speech}.}, and a smaller
proportion (approximately $8\%$ of the training set) of internal audio,
including synthetic text-to-speech clone recordings of the LJSpeech transcripts, spoken-digit
and telephony-style call recordings, and other internally collected speech data not
publicly available. The exact file count contributed by
each source is given in \autoref{tab:trainset} (\appref{appx:extra_tables}). In total the dataset
comprises $646,769$ files. Since the data points used to train the SAE are the individual time frames
from the Whisper encoder, this yields on the order of $200$ million
data points, more than sufficient to train the SAE's approximately 16 million parameters.

Additional data was used in some analyses to evaluate the trained model, these
datasets include Mozilla Common Voice Spanish and French.

\subsection{Training Details}
\label{sec:training_details}

The latent dimension (16,000) and TopK sparsity ($k=45$) were selected via a
preliminary grid search over $k \in \{5, 10, 20, 25, 50, 100\}$ and latent
dimension $\in \{1000, 2000, 4000, 8000, 10000, 16000\}$, evaluated on validation
reconstruction loss. Both larger $k$ and larger latent dimension
substantially improved reconstruction (e.g., $k{=}5$, dimension $1000$
reached a validation MSE of $0.232$, versus $0.038$ for $k{=}50$, dimension
$16{,}000$), though the relationship was not strictly monotonic at the
extremes (the single best cell in the grid, $k{=}100$, dimension
$10{,}000$, reached $0.033$, while $k{=}100$, dimension $16{,}000$ reached
only $0.049$). The final configuration was chosen within this strong-performing
region, favoring a lower $k$ than the grid's single best cell to encourage
sparser, more atomic activations at a modest cost to reconstruction fidelity,
consistent with sparsity/dimension choices reported elsewhere in the SAE
literature \citep{gao2025scaling}.

The SAE was trained with the Adam optimizer \citep{kingma2014adam} to convergence over the training set, using early stopping on a held-out validation split; \appref{appx:training_config} details the optimizer configuration and stopping criteria.

\subsection{Steering}
\label{sec:steering}
Steering is a causal intervention: rather than merely observing that a latent's activation correlates with a hypothesized feature, we directly manipulate that latent's value and observe the resulting effect on Whisper's output. Given a latent hypothesized to represent a specific feature, we test this in two complementary directions. In \textit{negative steering}, we set the latent's activation to a negative value in frames where it was previously active, suppressing the feature; in \textit{positive steering}, we set the latent's activation to a positive value in frames where it was previously inactive, inserting the feature where it was previously absent. In both cases, the modified sparse latent vector is passed through the SAE decoder to produce a modified reconstruction of the Whisper embedding (\autoref{fig:whisper_arch}), which is then passed to the Whisper decoder to generate a new transcript. If negative steering causes the hypothesized feature to disappear from the transcript, or positive steering causes the feature to appear where it was previously absent, this provides direct causal evidence that the latent encodes the feature rather than merely correlating with it, a substantially stronger form of evidence than correlational precision and recall alone.

An important methodological detail concerns \textit{where} the intervention is applied. Setting a latent's activation across every frame of an utterance produces a much stronger perturbation than the feature's natural footprint, which is typically confined to a small number of frames corresponding to a single word or phrase. In practice, whole-utterance intervention overwhelms the reconstruction, including the profanity and Numbers latents, reliably causing the Whisper decoder to collapse into repeating a single unrelated token or language tag rather than producing an interpretable change. We therefore use a frame-localized procedure as our default across every steering result reported in this paper: the latent's value is modified only within the frames spanned by a single word.

\subsection{Feature Identification}
\label{sec:feature_id}
We characterize the SAE's latent space using two complementary strategies: targeted, hypothesis-driven validation of specific candidate features, and a comprehensive automated labeling pass across the full latent space. For candidate features motivated by linguistic theory (for example, a specific word, a morphological suffix, or a semantic category such as profanity), we searched the latent space for latents whose activation correlated with the presence of that feature in our data. Precision and recall are computed at the level of individual occurrences (e.g., a single word or phone instance) throughout this paper, not distinct word or phone types. Ground-truth labels for whether a given feature was present in a file were manually or automatically annotated depending on the feature type; \appref{appx:labeling} describes these annotation procedures in detail. To characterize the latent space more broadly, beyond the specific latents examined this way, we supplement this analysis with a fully automated labeling procedure applied to every latent.

\subsubsection{Analysis Set}
\label{sec:analysis_set}
To support this analysis, a dedicated set of 28,414 audio files was constructed. The analysis set comprises 13,084 files from LJSpeech-1.1, 7,830 English and 7,500 Spanish files randomly sampled from Mozilla Common Voice (v13.0), and 331 additional clips from a curated profanity subset drawn from Mozilla Common Voice English. \autoref{tab:analysisset} (\appref{appx:extra_tables}) summarizes the composition. The analysis set partially overlaps with the SAE training data. This partial overlap is inconsequential regardless of its exact extent. The SAE is trained solely to reconstruct Whisper's own encoder output, and Whisper's encoder itself is frozen throughout, never fine-tuned on any data used in this work. The SAE therefore cannot learn structure that is not already present in Whisper's fixed representations, whether or not a given analysis file happened to also appear in the SAE's training set; a formal explanation of this claim is provided in \appref{appx:data_trans}.

In the case of the phoneme, morpheme, and lexical analyses, the dataset was subset to use the LJSpeech corpus. LJSpeech has TextGrids that align the audio to phone-level transcriptions, which enables the examination of phonological sequences that may be spelt differently.

For all other analyses, we used the entirety of the test set. In those cases, alignment was performed using a forced-alignment tool to produce character- and word-level alignments for each file. For consistency, this same forced-alignment procedure was also applied to the LJSpeech corpus in these analyses.

\subsubsection{Automated Labeling}
\label{sec:auto-label}
To supplement the manual feature analysis, all latents were labeled using GPT-OSS 120B into one of seven categories (phonetic, orthographic, morphological, lexical, semantic, syntactic, or diffuse), validated against a manually-evaluated subset with an accuracy of 76\%; \appref{appx:auto_label} describes the labeling procedure in detail. These results are reported in \appref{appx:latent_samples}.

\section{Whisper SAE Results} \label{sec:whisper}

This section presents the results of the feature analyses. We find that Whisper encodes a surprisingly rich hierarchy of representations, spanning low-level acoustic properties, positional information, and abstract linguistic features up to the level of semantics. However, latent indices exhibit complex
behavior with both splitting and absorption: a latent associated with a feature
may not be activated for all samples containing that feature, and there may be other
latents with more narrow feature characteristics \citep{chanin2025a}.

We present our results organized by feature type: phonetics, lexical items, syntax and morphology, and finally semantics. The focus is entirely on linguistic information; however, there is evidence that non-linguistic information is also contained within the latent space, see \appref{appx:others}.

\autoref{tab:overview} summarizes the results presented in this section: for each level of the hierarchy of linguistic abstraction motivated in \autoref{sec:intro}, phonetic and phonological structure, lexical identity, syntactic and morphological structure, and semantic content, we present at least one representative latent along with its precision and recall for the hypothesized feature. A language-identity latent is discussed separately in \appref{appx:extra_tables}. These examples are illustrative rather than exhaustive: all latents were systematically labeled into one of seven categories using an automated pipeline, and \appref{appx:latent_samples} reports a broader sample of these labeled latents beyond the specific cases examined in detail below.

\begin{table}[htbp]
\centering
\footnotesize
\caption{Overview of representative latents discussed in this section, organized by level of the linguistic hierarchy examined. These are illustrative examples rather than an exhaustive account; \appref{appx:latent_samples} reports a broader sample of automatically labeled latents.}
\label{tab:overview}
\begin{tabular}{llrr}
\toprule
\textbf{Level} & \textbf{Latent ID} & \textbf{Precision} & \textbf{Recall} \\
\midrule
Phonetic & 6373 & 87.8\% & 68.4\% \\
Lexical & 1445 & 86.9\% & 100\% \\
Syntactic & 1163 & 74.8\% & 41.5\% \\
Morphological & 29 & 78.4\% & 15.7\% \\
Semantic (Numbers) & 1492 & 93.5\% & 37.7\% \\
Semantic (Profanity) & 3584, 104 & 3.9\% & 89.9\% \\
\bottomrule
\end{tabular}
\end{table}

\subsection{Phonetic Features}
\label{sec:phonetic}

\begin{figure}[htbp] \centering
    \includegraphics[width=0.5\linewidth]{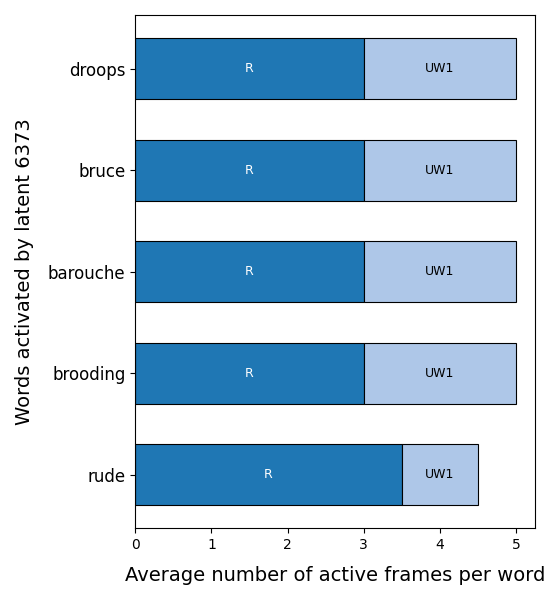}
    \caption{The top 5 words with the highest average number of frames with \li{6373}
        activated (precision value of $87.8\%$, recall value of $68.4\%$). The different colored bars represent the average amount of activated frames that each phone in the word received. Phones absent in plot were not activated for this latent.}
    \label{fig:phone_ruw1}
\end{figure}

We first present evidence for phonetic features, which we consider to be among the least surprising of the linguistic information encoded by Whisper since they seem the most necessary for the task of transcription. Despite this necessity, however, phonetic representations are themselves not trivially separable into individual features because the acoustic properties of phones vary depending on neighboring phones due to co-articulation \citep{daniloff1973defining}. Nonetheless, the Whisper encoder shows clear evidence of representing phonetic information. \autoref{fig:phone_ruw1} (\appref{appx:latent_samples})
demonstrates an example of a phonetic latent representing the diphone R UW1 (precision value of $87.8\%$, recall value of $68.4\%$). Many further
examples can be found in \appref{appx:latent_samples}. We attempted causal steering of \li{6373} in both directions: negative steering on genuine R-UW1 occurrences, and positive steering on words chosen so that inserting or substituting in R-UW1 could plausibly form a real word (for example \textit{in}$\rightarrow$\textit{ruin}, \textit{food}$\rightarrow$\textit{rude}). Neither direction produced a clean substitution. This was the least steerable latent we examined (\autoref{sec:discussions}).

\subsection{Lexical Features}
\label{sec:words}

The existence of phonetic features is intuitive given that the system's goal is
to transcribe speech. However, lexical features also seem possible especially
in English where there is not a 1-to-1 mapping between orthographic
representations and phonetic representations. For example, the letter ``a'' in (American) English can represent several different vowels, such as \textipa{/æ/} (as in \textit{cat}), \textipa{/A/} (as in \textit{father}), \textipa{/eI/} (as in \textit{cake}), \textipa{/E/} (as in \textit{many}), and \textipa{@} (as in \textit{about}). As a result, it may be advantageous for Whisper to encode lexical information in addition to phonetic information. Indeed, this seems to be the case, as we see features that do seem to be convincingly lexical such as \li{1445}, which represents the word \textit{nothing} and has a precision
value of $86.9\%$ and a recall value of $100\%$. Positive steering of \li{1445} results in the insertion of the word \textit{nothing} while leaving the rest of the sentence intact, e.g.,\ ``At the present day, writes Huxley.'' becomes ``At the present, nothing writes Huxley.''; negative steering failed to remove the word \textit{nothing}.

\subsection{Syntactic and Morphological Features}
\label{sec:syntactic_and_morpho}
Beyond individual lexical items, we also find evidence that Whisper represents
syntactic and morphological latents, rather than only single words. \li{1163}
activates on third-person pronouns with a precision of $74.8\%$, far above
chance, and a recall of $41.5\%$. Critically, this activation is not concentrated on a
single pronoun: of the latent's activations that fall within this word set,
$66.9\%$ are `he', $16.2\%$ are `they', and $9.8\%$ are `she', with the
remainder split across the less frequent third-person forms. This
distinguishes the latent from a purely lexical detector for `he' specifically.

Positive steering of \li{1163} reliably inserts a third-person pronoun (\textit{he}). As with \li{1445}, negative steering of \li{1163} failed to remove the pronoun.

\begin{figure}[htbp] \centering
    \includegraphics[width=0.5\linewidth]{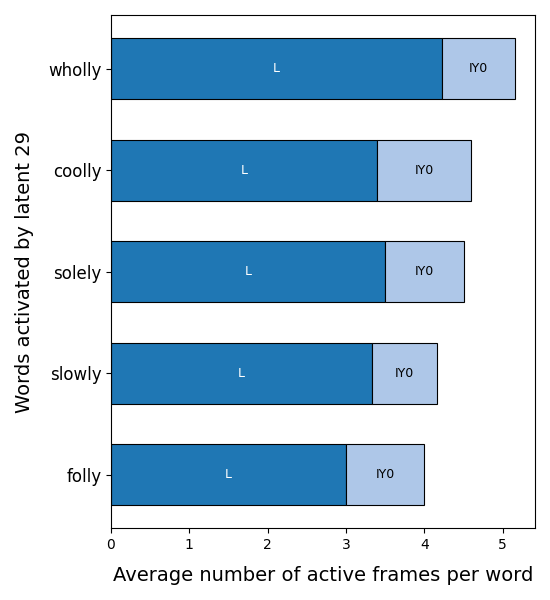}
    \caption{The top 5 words with the highest average number of frames with \li{29}
        activated (precision value of $78.4\%$, recall value of $15.7\%$). The different colored bars represent the average amount of activated frames that each phone in the word received. Phones absent in plot were not activated for this latent.}
    \label{fig:morpheme_ly}
\end{figure}

Morphological information is also not strictly necessary to transcribe audio.
However, given that the sparse autoencoder already appears to encode individual words, the model may also encode other
abstract representations. Examination of the SAE latents confirms
that Whisper learns morphological representations.
\autoref{fig:morpheme_ly} (\appref{appx:latent_samples}) demonstrates the words which have the largest ratio of
activated frames for \li{29}. This latent index correlates with the suffix `ly', which in English serves as an adverbial marker
(precision value of $78.4\%$, recall value of $15.7\%$). While the recall value is somewhat low for this latent, this is likely due to feature splitting: there are multiple different `ly' adverbial latents, such as latent 9779 (see \appref{appx:latent_samples} for an example). Positive steering of \li{29} on adjectives capable of taking the `-ly' suffix produced a clean, correctly-suffixed adverb, e.g.,\ \textit{particular}$\rightarrow$\textit{particularly}, \textit{certain}$\rightarrow$\textit{certainly}. Negative steering showed no effect.

\subsection{Semantic Features}
\label{sec:semantic}

Despite not being strictly necessary for transcription, semantic representations may help resolve ambiguities such as homophones. Even without homophones, noise or other interference can degrade transcription in ways that contextual understanding can help resolve. Notably, we find that such semantic representations are indeed present in the Whisper encoder. Two types of semantic features are presented: numbers and profanity.

\subsubsection{Numbers}
One prevalent type of semantic feature in the dataset is numerical
values. \Li{1492} has a strong
association with many different numbers, with a precision of $93.5\%$ and a recall of $37.7\%$; \autoref{tab:numbers} shows its activation on four such samples. Its activations are concentrated on values such as `twenty', `sixty', `thirty', `fifty', and `forty', but the latent is not a detector for any single one of these: it activates across at least 21 distinct number words. The comparatively low recall is likely impacted by feature splitting: many individual numbers, particularly the most frequent ones such as `one' and `two', appear to have their own dedicated latent features.

We validated this correlational finding causally using positive steering on \li{1492}. This produced a clean causal effect, replacing a single word with a number while leaving the rest of the sentence completely intact: ``constantly in the Ascendant'' became ``constantly in the 80's''; ``the composition of our federal courts'' became ``the composition of our 50 crores''\footnote{Interestingly, \textit{Crore} is a Hindi word which also has a meaning that is numeric in nature (specifically it is a unit of measurement meaning ten million).}; and ``the governor of Newgate'' became ``the governor of 90-80''. Negative steering produced no consistent effect, however, suggesting this latent may be somewhat weaker than the profanity latent.

\begin{table}[htbp]
\footnotesize
\centering
\caption{Activation of \li{1492} for four samples containing numbers. The darker
highlighting shows a stronger activation value (normalized per sample).}
\label{tab:numbers}
\begin{tabular}{|l|}
\hline
{\setlength{\fboxsep}{0pt}%
\colorbox[rgb]{1.000,1.000,1.000}{\strut\textcolor{black}{f}}%
\colorbox[rgb]{1.000,1.000,1.000}{\strut\textcolor{black}{o}}%
\colorbox[rgb]{1.000,1.000,1.000}{\strut\textcolor{black}{u}}%
\colorbox[rgb]{1.000,1.000,1.000}{\strut\textcolor{black}{r}}%
\colorbox[rgb]{1.000,1.000,1.000}{\strut\textcolor{black}{t}}%
\colorbox[rgb]{1.000,1.000,1.000}{\strut\textcolor{black}{e}}%
\colorbox[rgb]{1.000,1.000,1.000}{\strut\textcolor{black}{e}}%
\colorbox[rgb]{1.000,1.000,1.000}{\strut\textcolor{black}{n}}%
\colorbox[rgb]{1.000,1.000,1.000}{\strut\kern0.35em}%
\colorbox[rgb]{1.000,0.647,0.000}{\strut\textcolor{white}{s}}%
\colorbox[rgb]{1.000,0.647,0.000}{\strut\textcolor{white}{i}}%
\colorbox[rgb]{1.000,0.647,0.000}{\strut\textcolor{white}{x}}%
\colorbox[rgb]{1.000,0.647,0.000}{\strut\textcolor{white}{t}}%
\colorbox[rgb]{1.000,0.647,0.000}{\strut\textcolor{white}{y}}%
\colorbox[rgb]{1.000,1.000,1.000}{\strut\kern0.35em}%
\colorbox[rgb]{1.000,1.000,1.000}{\strut\textcolor{black}{n}}%
\colorbox[rgb]{1.000,1.000,1.000}{\strut\textcolor{black}{i}}%
\colorbox[rgb]{1.000,1.000,1.000}{\strut\textcolor{black}{n}}%
\colorbox[rgb]{1.000,1.000,1.000}{\strut\textcolor{black}{e}}%
\colorbox[rgb]{1.000,1.000,1.000}{\strut\kern0.35em}%
\colorbox[rgb]{1.000,1.000,1.000}{\strut\textcolor{black}{f}}%
\colorbox[rgb]{1.000,1.000,1.000}{\strut\textcolor{black}{o}}%
\colorbox[rgb]{1.000,1.000,1.000}{\strut\textcolor{black}{u}}%
\colorbox[rgb]{1.000,1.000,1.000}{\strut\textcolor{black}{r}}%
\colorbox[rgb]{1.000,1.000,1.000}{\strut\textcolor{black}{t}}%
\colorbox[rgb]{1.000,1.000,1.000}{\strut\textcolor{black}{e}}%
\colorbox[rgb]{1.000,1.000,1.000}{\strut\textcolor{black}{e}}%
\colorbox[rgb]{1.000,1.000,1.000}{\strut\textcolor{black}{n}}%
\colorbox[rgb]{1.000,1.000,1.000}{\strut\kern0.35em}%
\colorbox[rgb]{1.000,0.742,0.268}{\strut\textcolor{white}{s}}%
\colorbox[rgb]{1.000,0.742,0.268}{\strut\textcolor{white}{e}}%
\colorbox[rgb]{1.000,0.742,0.268}{\strut\textcolor{white}{v}}%
\colorbox[rgb]{1.000,0.742,0.268}{\strut\textcolor{white}{e}}%
\colorbox[rgb]{1.000,0.742,0.268}{\strut\textcolor{white}{n}}%
\colorbox[rgb]{1.000,0.742,0.268}{\strut\textcolor{white}{t}}%
\colorbox[rgb]{1.000,0.742,0.268}{\strut\textcolor{white}{y}}%
}\\%
\hline
{\setlength{\fboxsep}{0pt}%
\colorbox[rgb]{1.000,1.000,1.000}{\strut\textcolor{black}{o}}%
\colorbox[rgb]{1.000,1.000,1.000}{\strut\textcolor{black}{n}}%
\colorbox[rgb]{1.000,1.000,1.000}{\strut\textcolor{black}{e}}%
\colorbox[rgb]{1.000,1.000,1.000}{\strut\kern0.35em}%
\colorbox[rgb]{1.000,0.668,0.060}{\strut\textcolor{white}{s}}%
\colorbox[rgb]{1.000,0.668,0.060}{\strut\textcolor{white}{i}}%
\colorbox[rgb]{1.000,0.668,0.060}{\strut\textcolor{white}{x}}%
\colorbox[rgb]{1.000,0.668,0.060}{\strut\textcolor{white}{t}}%
\colorbox[rgb]{1.000,0.668,0.060}{\strut\textcolor{white}{y}}%
\colorbox[rgb]{1.000,1.000,1.000}{\strut\kern0.35em}%
\colorbox[rgb]{1.000,1.000,1.000}{\strut\textcolor{black}{t}}%
\colorbox[rgb]{1.000,1.000,1.000}{\strut\textcolor{black}{o}}%
\colorbox[rgb]{1.000,1.000,1.000}{\strut\kern0.35em}%
\colorbox[rgb]{1.000,1.000,1.000}{\strut\textcolor{black}{o}}%
\colorbox[rgb]{1.000,1.000,1.000}{\strut\textcolor{black}{n}}%
\colorbox[rgb]{1.000,1.000,1.000}{\strut\textcolor{black}{e}}%
\colorbox[rgb]{1.000,1.000,1.000}{\strut\kern0.35em}%
\colorbox[rgb]{1.000,0.647,0.000}{\strut\textcolor{white}{s}}%
\colorbox[rgb]{1.000,0.647,0.000}{\strut\textcolor{white}{e}}%
\colorbox[rgb]{1.000,0.647,0.000}{\strut\textcolor{white}{v}}%
\colorbox[rgb]{1.000,0.647,0.000}{\strut\textcolor{white}{e}}%
\colorbox[rgb]{1.000,0.647,0.000}{\strut\textcolor{white}{n}}%
\colorbox[rgb]{1.000,0.647,0.000}{\strut\textcolor{white}{t}}%
\colorbox[rgb]{1.000,0.647,0.000}{\strut\textcolor{white}{y}}%
\colorbox[rgb]{1.000,1.000,1.000}{\strut\kern0.35em}%
\colorbox[rgb]{1.000,1.000,1.000}{\strut\textcolor{black}{p}}%
\colorbox[rgb]{1.000,1.000,1.000}{\strut\textcolor{black}{o}}%
\colorbox[rgb]{1.000,1.000,1.000}{\strut\textcolor{black}{u}}%
\colorbox[rgb]{1.000,1.000,1.000}{\strut\textcolor{black}{n}}%
\colorbox[rgb]{1.000,1.000,1.000}{\strut\textcolor{black}{d}}%
\colorbox[rgb]{1.000,1.000,1.000}{\strut\textcolor{black}{s}}%
}\\%
\hline
{\setlength{\fboxsep}{0pt}%
\colorbox[rgb]{1.000,1.000,1.000}{\strut\textcolor{black}{t}}%
\colorbox[rgb]{1.000,1.000,1.000}{\strut\textcolor{black}{o}}%
\colorbox[rgb]{1.000,1.000,1.000}{\strut\kern0.35em}%
\colorbox[rgb]{1.000,0.665,0.050}{\strut\textcolor{white}{f}}%
\colorbox[rgb]{1.000,0.665,0.050}{\strut\textcolor{white}{i}}%
\colorbox[rgb]{1.000,0.665,0.050}{\strut\textcolor{white}{f}}%
\colorbox[rgb]{1.000,0.665,0.050}{\strut\textcolor{white}{t}}%
\colorbox[rgb]{1.000,0.665,0.050}{\strut\textcolor{white}{y}}%
\colorbox[rgb]{1.000,1.000,1.000}{\strut\kern0.35em}%
\colorbox[rgb]{1.000,1.000,1.000}{\strut\textcolor{black}{s}}%
\colorbox[rgb]{1.000,1.000,1.000}{\strut\textcolor{black}{i}}%
\colorbox[rgb]{1.000,1.000,1.000}{\strut\textcolor{black}{x}}%
\colorbox[rgb]{1.000,1.000,1.000}{\strut\kern0.35em}%
\colorbox[rgb]{1.000,1.000,1.000}{\strut\textcolor{black}{i}}%
\colorbox[rgb]{1.000,1.000,1.000}{\strut\textcolor{black}{n}}%
\colorbox[rgb]{1.000,1.000,1.000}{\strut\kern0.35em}%
\colorbox[rgb]{1.000,1.000,1.000}{\strut\textcolor{black}{e}}%
\colorbox[rgb]{1.000,1.000,1.000}{\strut\textcolor{black}{i}}%
\colorbox[rgb]{1.000,1.000,1.000}{\strut\textcolor{black}{g}}%
\colorbox[rgb]{1.000,1.000,1.000}{\strut\textcolor{black}{h}}%
\colorbox[rgb]{1.000,1.000,1.000}{\strut\textcolor{black}{t}}%
\colorbox[rgb]{1.000,1.000,1.000}{\strut\textcolor{black}{e}}%
\colorbox[rgb]{1.000,1.000,1.000}{\strut\textcolor{black}{e}}%
\colorbox[rgb]{1.000,1.000,1.000}{\strut\textcolor{black}{n}}%
\colorbox[rgb]{1.000,1.000,1.000}{\strut\kern0.35em}%
\colorbox[rgb]{1.000,0.647,0.000}{\strut\textcolor{white}{t}}%
\colorbox[rgb]{1.000,0.647,0.000}{\strut\textcolor{white}{h}}%
\colorbox[rgb]{1.000,0.647,0.000}{\strut\textcolor{white}{i}}%
\colorbox[rgb]{1.000,0.647,0.000}{\strut\textcolor{white}{r}}%
\colorbox[rgb]{1.000,0.647,0.000}{\strut\textcolor{white}{t}}%
\colorbox[rgb]{1.000,0.647,0.000}{\strut\textcolor{white}{y}}%
\colorbox[rgb]{1.000,1.000,1.000}{\strut\kern0.35em}%
\colorbox[rgb]{1.000,1.000,1.000}{\strut\textcolor{black}{n}}%
\colorbox[rgb]{1.000,1.000,1.000}{\strut\textcolor{black}{i}}%
\colorbox[rgb]{1.000,1.000,1.000}{\strut\textcolor{black}{n}}%
\colorbox[rgb]{1.000,1.000,1.000}{\strut\textcolor{black}{e}}%
}\\%
\hline
{\setlength{\fboxsep}{0pt}%
\colorbox[rgb]{1.000,1.000,1.000}{\strut\textcolor{black}{w}}%
\colorbox[rgb]{1.000,1.000,1.000}{\strut\textcolor{black}{a}}%
\colorbox[rgb]{1.000,1.000,1.000}{\strut\textcolor{black}{s}}%
\colorbox[rgb]{1.000,1.000,1.000}{\strut\kern0.35em}%
\colorbox[rgb]{1.000,1.000,1.000}{\strut\textcolor{black}{o}}%
\colorbox[rgb]{1.000,1.000,1.000}{\strut\textcolor{black}{n}}%
\colorbox[rgb]{1.000,1.000,1.000}{\strut\kern0.35em}%
\colorbox[rgb]{1.000,1.000,1.000}{\strut\textcolor{black}{n}}%
\colorbox[rgb]{1.000,1.000,1.000}{\strut\textcolor{black}{o}}%
\colorbox[rgb]{1.000,1.000,1.000}{\strut\textcolor{black}{v}}%
\colorbox[rgb]{1.000,1.000,1.000}{\strut\textcolor{black}{e}}%
\colorbox[rgb]{1.000,1.000,1.000}{\strut\textcolor{black}{m}}%
\colorbox[rgb]{1.000,1.000,1.000}{\strut\textcolor{black}{b}}%
\colorbox[rgb]{1.000,1.000,1.000}{\strut\textcolor{black}{e}}%
\colorbox[rgb]{1.000,1.000,1.000}{\strut\textcolor{black}{r}}%
\colorbox[rgb]{1.000,1.000,1.000}{\strut\kern0.35em}%
\colorbox[rgb]{1.000,1.000,1.000}{\strut\textcolor{black}{t}}%
\colorbox[rgb]{1.000,1.000,1.000}{\strut\textcolor{black}{w}}%
\colorbox[rgb]{1.000,1.000,1.000}{\strut\textcolor{black}{e}}%
\colorbox[rgb]{1.000,1.000,1.000}{\strut\textcolor{black}{n}}%
\colorbox[rgb]{1.000,1.000,1.000}{\strut\textcolor{black}{t}}%
\colorbox[rgb]{1.000,1.000,1.000}{\strut\textcolor{black}{y}}%
\colorbox[rgb]{1.000,1.000,1.000}{\strut\kern0.35em}%
\colorbox[rgb]{1.000,1.000,1.000}{\strut\textcolor{black}{t}}%
\colorbox[rgb]{1.000,1.000,1.000}{\strut\textcolor{black}{o}}%
\colorbox[rgb]{1.000,1.000,1.000}{\strut\kern0.35em}%
\colorbox[rgb]{1.000,0.839,0.545}{\strut\textcolor{black}{t}}%
\colorbox[rgb]{1.000,0.839,0.545}{\strut\textcolor{black}{w}}%
\colorbox[rgb]{1.000,0.839,0.545}{\strut\textcolor{black}{e}}%
\colorbox[rgb]{1.000,0.839,0.545}{\strut\textcolor{black}{n}}%
\colorbox[rgb]{1.000,0.839,0.545}{\strut\textcolor{black}{t}}%
\colorbox[rgb]{1.000,0.839,0.545}{\strut\textcolor{black}{y}}%
\colorbox[rgb]{1.000,1.000,1.000}{\strut\kern0.35em}%
\colorbox[rgb]{1.000,1.000,1.000}{\strut\textcolor{black}{o}}%
\colorbox[rgb]{1.000,1.000,1.000}{\strut\textcolor{black}{n}}%
\colorbox[rgb]{1.000,1.000,1.000}{\strut\textcolor{black}{e}}%
\colorbox[rgb]{1.000,1.000,1.000}{\strut\kern0.35em}%
\colorbox[rgb]{1.000,1.000,1.000}{\strut\textcolor{black}{n}}%
\colorbox[rgb]{1.000,1.000,1.000}{\strut\textcolor{black}{i}}%
\colorbox[rgb]{1.000,1.000,1.000}{\strut\textcolor{black}{n}}%
\colorbox[rgb]{1.000,1.000,1.000}{\strut\textcolor{black}{e}}%
\colorbox[rgb]{1.000,1.000,1.000}{\strut\textcolor{black}{t}}%
\colorbox[rgb]{1.000,1.000,1.000}{\strut\textcolor{black}{e}}%
\colorbox[rgb]{1.000,1.000,1.000}{\strut\textcolor{black}{e}}%
\colorbox[rgb]{1.000,1.000,1.000}{\strut\textcolor{black}{n}}%
\colorbox[rgb]{1.000,1.000,1.000}{\strut\kern0.35em}%
\colorbox[rgb]{1.000,0.647,0.000}{\strut\textcolor{white}{s}}%
\colorbox[rgb]{1.000,0.647,0.000}{\strut\textcolor{white}{i}}%
\colorbox[rgb]{1.000,0.647,0.000}{\strut\textcolor{white}{x}}%
\colorbox[rgb]{1.000,0.647,0.000}{\strut\textcolor{white}{t}}%
\colorbox[rgb]{1.000,0.647,0.000}{\strut\textcolor{white}{y}}%
\colorbox[rgb]{1.000,1.000,1.000}{\strut\kern0.35em}%
\colorbox[rgb]{1.000,1.000,1.000}{\strut\textcolor{black}{t}}%
\colorbox[rgb]{1.000,1.000,1.000}{\strut\textcolor{black}{h}}%
\colorbox[rgb]{1.000,1.000,1.000}{\strut\textcolor{black}{r}}%
\colorbox[rgb]{1.000,1.000,1.000}{\strut\textcolor{black}{e}}%
\colorbox[rgb]{1.000,1.000,1.000}{\strut\textcolor{black}{e}}%
}\\%
\hline
\end{tabular}
\end{table}

\subsubsection{Profanity}
\label{sec:profanity}

Beyond numerical values, we find that the Whisper encoder also represents a second, quite different category of semantic feature: profanity. Curse words are used in a wide variety of ways, including as intensifiers, and identifying a dedicated latent for them lets us test whether the causal steering results we observed for numbers generalize to other kinds of semantic content. In order to identify a profanity feature, we use the set of sentences containing profanity described earlier.
Examining the encodings of these audio samples showed that \li{3584} was correlated with profanity. Further examination showed that a significant number of utterances containing the word 'damn' were better represented using a separate latent, \li{104}. \autoref{tab:profanity} shows a sample of utterances which have activations of \li{3584} and \li{104}. Together these latents account for a recall of $89.9\%$ with a precision of $3.9\%$. Profanity is quite rare in this dataset, occurring in $0.11\%$ of the entire corpus based on a word-level count. Many of the false positives occur in utterances that do contain profanity, suggesting that the activation of these semantic latents may not be strongly temporally constrained, with some utterances having activations through many frames of the encoding.

\begin{table}[htbp]
\footnotesize
\centering
\caption{Activation of \li{3584} and \li{104} for samples containing profanity.}
\label{tab:profanity}
\begin{tabular}{|l|}
\hline
{\setlength{\fboxsep}{0pt}%
\colorbox[rgb]{1.000,1.000,1.000}{\strut\textcolor{black}{A}}%
\colorbox[rgb]{1.000,1.000,1.000}{\strut\textcolor{black}{l}}%
\colorbox[rgb]{1.000,1.000,1.000}{\strut\textcolor{black}{i}}%
\colorbox[rgb]{1.000,1.000,1.000}{\strut\textcolor{black}{c}}%
\colorbox[rgb]{1.000,1.000,1.000}{\strut\textcolor{black}{e}}%
\colorbox[rgb]{1.000,1.000,1.000}{\strut\kern0.35em}%
\colorbox[rgb]{1.000,1.000,1.000}{\strut\textcolor{black}{w}}%
\colorbox[rgb]{1.000,1.000,1.000}{\strut\textcolor{black}{h}}%
\colorbox[rgb]{1.000,1.000,1.000}{\strut\textcolor{black}{o}}%
\colorbox[rgb]{1.000,1.000,1.000}{\strut\kern0.35em}%
\colorbox[rgb]{1.000,1.000,1.000}{\strut\textcolor{black}{t}}%
\colorbox[rgb]{1.000,1.000,1.000}{\strut\textcolor{black}{h}}%
\colorbox[rgb]{1.000,1.000,1.000}{\strut\textcolor{black}{e}}%
\colorbox[rgb]{1.000,0.694,0.134}{\strut\kern0.35em}%
\colorbox[rgb]{1.000,0.647,0.000}{\strut\textcolor{white}{f}}%
\colorbox[rgb]{1.000,0.647,0.000}{\strut\textcolor{white}{u}}%
\colorbox[rgb]{1.000,1.000,1.000}{\strut\textcolor{black}{c}}%
\colorbox[rgb]{1.000,1.000,1.000}{\strut\textcolor{black}{k}}%
\colorbox[rgb]{1.000,1.000,1.000}{\strut\kern0.35em}%
\colorbox[rgb]{1.000,1.000,1.000}{\strut\textcolor{black}{i}}%
\colorbox[rgb]{1.000,1.000,1.000}{\strut\textcolor{black}{s}}%
\colorbox[rgb]{1.000,1.000,1.000}{\strut\kern0.35em}%
\colorbox[rgb]{1.000,1.000,1.000}{\strut\textcolor{black}{B}}%
\colorbox[rgb]{1.000,1.000,1.000}{\strut\textcolor{black}{o}}%
\colorbox[rgb]{1.000,1.000,1.000}{\strut\textcolor{black}{b}}%
}\\%

\hline
{\setlength{\fboxsep}{0pt}%
\colorbox[rgb]{1.000,1.000,1.000}{\strut\textcolor{black}{Y}}%
\colorbox[rgb]{1.000,1.000,1.000}{\strut\textcolor{black}{e}}%
\colorbox[rgb]{1.000,1.000,1.000}{\strut\textcolor{black}{s}}%
\colorbox[rgb]{1.000,1.000,1.000}{\strut\kern0.35em}%
\colorbox[rgb]{1.000,0.647,0.000}{\strut\textcolor{white}{f}}%
\colorbox[rgb]{1.000,0.801,0.437}{\strut\textcolor{white}{u}}%
\colorbox[rgb]{1.000,1.000,1.000}{\strut\textcolor{black}{c}}%
\colorbox[rgb]{1.000,1.000,1.000}{\strut\textcolor{black}{k}}%
}\\%

\hline
{\setlength{\fboxsep}{0pt}%
\colorbox[rgb]{1.000,1.000,1.000}{\strut\textcolor{black}{T}}%
\colorbox[rgb]{1.000,1.000,1.000}{\strut\textcolor{black}{h}}%
\colorbox[rgb]{1.000,1.000,1.000}{\strut\textcolor{black}{a}}%
\colorbox[rgb]{1.000,1.000,1.000}{\strut\textcolor{black}{t}}%
\colorbox[rgb]{1.000,1.000,1.000}{\strut\kern0.35em}%
\colorbox[rgb]{1.000,0.740,0.264}{\strut\textcolor{white}{d}}%
\colorbox[rgb]{1.000,1.000,1.000}{\strut\textcolor{black}{a}}%
\colorbox[rgb]{1.000,1.000,1.000}{\strut\textcolor{black}{m}}%
\colorbox[rgb]{1.000,1.000,1.000}{\strut\textcolor{black}{n}}%
\colorbox[rgb]{1.000,1.000,1.000}{\strut\kern0.35em}%
\colorbox[rgb]{1.000,1.000,1.000}{\strut\textcolor{black}{m}}%
\colorbox[rgb]{1.000,1.000,1.000}{\strut\textcolor{black}{o}}%
\colorbox[rgb]{1.000,1.000,1.000}{\strut\textcolor{black}{u}}%
\colorbox[rgb]{1.000,1.000,1.000}{\strut\textcolor{black}{s}}%
\colorbox[rgb]{1.000,1.000,1.000}{\strut\textcolor{black}{e}}%
\colorbox[rgb]{1.000,1.000,1.000}{\strut\kern0.35em}%
\colorbox[rgb]{1.000,1.000,1.000}{\strut\textcolor{black}{c}}%
\colorbox[rgb]{1.000,1.000,1.000}{\strut\textcolor{black}{o}}%
\colorbox[rgb]{1.000,1.000,1.000}{\strut\textcolor{black}{u}}%
\colorbox[rgb]{1.000,1.000,1.000}{\strut\textcolor{black}{l}}%
\colorbox[rgb]{1.000,1.000,1.000}{\strut\textcolor{black}{d}}%
\colorbox[rgb]{1.000,1.000,1.000}{\strut\kern0.35em}%
\colorbox[rgb]{1.000,1.000,1.000}{\strut\textcolor{black}{s}}%
\colorbox[rgb]{1.000,1.000,1.000}{\strut\textcolor{black}{e}}%
\colorbox[rgb]{1.000,1.000,1.000}{\strut\textcolor{black}{n}}%
\colorbox[rgb]{1.000,1.000,1.000}{\strut\textcolor{black}{s}}%
\colorbox[rgb]{1.000,1.000,1.000}{\strut\textcolor{black}{e}}%
\colorbox[rgb]{1.000,1.000,1.000}{\strut\kern0.35em}%
\colorbox[rgb]{1.000,1.000,1.000}{\strut\textcolor{black}{m}}%
\colorbox[rgb]{1.000,1.000,1.000}{\strut\textcolor{black}{y}}%
\colorbox[rgb]{1.000,1.000,1.000}{\strut\kern0.35em}%
\colorbox[rgb]{1.000,1.000,1.000}{\strut\textcolor{black}{a}}%
\colorbox[rgb]{1.000,1.000,1.000}{\strut\textcolor{black}{n}}%
\colorbox[rgb]{1.000,1.000,1.000}{\strut\textcolor{black}{g}}%
\colorbox[rgb]{1.000,1.000,1.000}{\strut\textcolor{black}{e}}%
\colorbox[rgb]{1.000,1.000,1.000}{\strut\textcolor{black}{r}}%
}\\%

\hline
{\setlength{\fboxsep}{0pt}%
\colorbox[rgb]{1.000,0.647,0.000}{\strut\textcolor{white}{F}}%
\colorbox[rgb]{1.000,0.647,0.000}{\strut\textcolor{white}{u}}%
\colorbox[rgb]{1.000,1.000,1.000}{\strut\textcolor{black}{c}}%
\colorbox[rgb]{1.000,1.000,1.000}{\strut\textcolor{black}{k}}%
\colorbox[rgb]{1.000,1.000,1.000}{\strut\kern0.35em}%
\colorbox[rgb]{1.000,1.000,1.000}{\strut\textcolor{black}{R}}%
\colorbox[rgb]{1.000,1.000,1.000}{\strut\textcolor{black}{i}}%
\colorbox[rgb]{1.000,1.000,1.000}{\strut\textcolor{black}{c}}%
\colorbox[rgb]{1.000,1.000,1.000}{\strut\textcolor{black}{a}}%
\colorbox[rgb]{1.000,1.000,1.000}{\strut\textcolor{black}{r}}%
\colorbox[rgb]{1.000,1.000,1.000}{\strut\textcolor{black}{d}}%
\colorbox[rgb]{1.000,1.000,1.000}{\strut\textcolor{black}{o}}%
\colorbox[rgb]{1.000,1.000,1.000}{\strut\kern0.35em}%
\colorbox[rgb]{1.000,1.000,1.000}{\strut\textcolor{black}{i}}%
\colorbox[rgb]{1.000,1.000,1.000}{\strut\textcolor{black}{t}}%
\colorbox[rgb]{1.000,1.000,1.000}{\strut\textcolor{black}{'}}%
\colorbox[rgb]{1.000,1.000,1.000}{\strut\textcolor{black}{s}}%
\colorbox[rgb]{1.000,1.000,1.000}{\strut\kern0.35em}%
\colorbox[rgb]{1.000,1.000,1.000}{\strut\textcolor{black}{j}}%
\colorbox[rgb]{1.000,1.000,1.000}{\strut\textcolor{black}{u}}%
\colorbox[rgb]{1.000,1.000,1.000}{\strut\textcolor{black}{s}}%
\colorbox[rgb]{1.000,1.000,1.000}{\strut\textcolor{black}{t}}%
\colorbox[rgb]{1.000,1.000,1.000}{\strut\kern0.35em}%
\colorbox[rgb]{1.000,1.000,1.000}{\strut\textcolor{black}{t}}%
\colorbox[rgb]{1.000,1.000,1.000}{\strut\textcolor{black}{h}}%
\colorbox[rgb]{1.000,1.000,1.000}{\strut\textcolor{black}{a}}%
\colorbox[rgb]{1.000,1.000,1.000}{\strut\textcolor{black}{t}}%
}\\%

\hline
{\setlength{\fboxsep}{0pt}%
\colorbox[rgb]{1.000,1.000,1.000}{\strut\textcolor{black}{T}}%
\colorbox[rgb]{1.000,1.000,1.000}{\strut\textcolor{black}{o}}%
\colorbox[rgb]{1.000,1.000,1.000}{\strut\textcolor{black}{m}}%
\colorbox[rgb]{1.000,1.000,1.000}{\strut\kern0.35em}%
\colorbox[rgb]{1.000,1.000,1.000}{\strut\textcolor{black}{c}}%
\colorbox[rgb]{1.000,1.000,1.000}{\strut\textcolor{black}{a}}%
\colorbox[rgb]{1.000,1.000,1.000}{\strut\textcolor{black}{n}}%
\colorbox[rgb]{1.000,1.000,1.000}{\strut\textcolor{black}{'}}%
\colorbox[rgb]{1.000,1.000,1.000}{\strut\textcolor{black}{t}}%
\colorbox[rgb]{1.000,1.000,1.000}{\strut\kern0.35em}%
\colorbox[rgb]{1.000,1.000,1.000}{\strut\textcolor{black}{k}}%
\colorbox[rgb]{1.000,1.000,1.000}{\strut\textcolor{black}{e}}%
\colorbox[rgb]{1.000,1.000,1.000}{\strut\textcolor{black}{e}}%
\colorbox[rgb]{1.000,1.000,1.000}{\strut\textcolor{black}{p}}%
\colorbox[rgb]{1.000,1.000,1.000}{\strut\kern0.35em}%
\colorbox[rgb]{1.000,1.000,1.000}{\strut\textcolor{black}{h}}%
\colorbox[rgb]{1.000,1.000,1.000}{\strut\textcolor{black}{i}}%
\colorbox[rgb]{1.000,1.000,1.000}{\strut\textcolor{black}{s}}%
\colorbox[rgb]{1.000,1.000,1.000}{\strut\kern0.35em}%
\colorbox[rgb]{1.000,0.647,0.000}{\strut\textcolor{white}{s}}%
\colorbox[rgb]{1.000,0.647,0.000}{\strut\textcolor{white}{h}}%
\colorbox[rgb]{1.000,1.000,1.000}{\strut\textcolor{black}{i}}%
\colorbox[rgb]{1.000,1.000,1.000}{\strut\textcolor{black}{t}}%
\colorbox[rgb]{1.000,1.000,1.000}{\strut\kern0.35em}%
\colorbox[rgb]{1.000,1.000,1.000}{\strut\textcolor{black}{t}}%
\colorbox[rgb]{1.000,1.000,1.000}{\strut\textcolor{black}{o}}%
\colorbox[rgb]{1.000,1.000,1.000}{\strut\textcolor{black}{g}}%
\colorbox[rgb]{1.000,1.000,1.000}{\strut\textcolor{black}{e}}%
\colorbox[rgb]{1.000,1.000,1.000}{\strut\textcolor{black}{t}}%
\colorbox[rgb]{1.000,1.000,1.000}{\strut\textcolor{black}{h}}%
\colorbox[rgb]{1.000,1.000,1.000}{\strut\textcolor{black}{e}}%
\colorbox[rgb]{1.000,1.000,1.000}{\strut\textcolor{black}{r}}%
}\\%

\hline
\end{tabular}
\end{table}

Profanity's precision ($3.9\%$) is substantially weaker than that of the Numbers feature ($93.5\%$), the weakest correlational evidence of any feature examined in this section. To confirm that \li{3584} is nonetheless causally linked to profanity rather than merely correlated with it, we validated it causally using steering. Negative steering reliably replaces the curse word with a phonetically similar word while leaving the rest of the sentence intact, as shown in \autoref{tab:profanity_steer}. Positive steering succeeds as well when applied to a word phonetically close to a curse word: positively steering \li{3584} turns \textit{sheet} into \textit{shit}, while leaving the rest of the sentence intact.

\begin{table*}[htbp]
\centering
\footnotesize
\caption{Effect of frame-localized negative steering (a negative multiplier applied only within the curse word's own frame span) on \li{3584}/\li{104}. The Original column is this model's own baseline transcript of the curated audio before any steering is applied.}
\label{tab:profanity_steer}
\begin{tabular}{|p{0.45\textwidth}|p{0.45\textwidth}|}
\hline
    \textbf{Original (baseline transcript)} & \textbf{Steered (curse word's frame span only)} \\
\hline
\hline
    at least who the \textbf{fuck} is Bob? & at least who's the focus Bob \\
\hline
    Yes, \textbf{fuck}. & Yes, Falk. \\
\hline
    That \textbf{damn} mouse could sense my anger. & That them mouse could sense my anger. \\
\hline
    I'm not talking about this \textbf{shit}. & I'm not talking about this yet. \\
\hline
\end{tabular}
\end{table*}

\subsubsection{Cross-Lingual Activation}

The results of \citet{templeton2024scaling} suggest that many latent indices may have
semantic activations that are independent of language. Since Whisper is trained
to be a multilingual ASR, it is possible to examine the same effect by curating audio clips with curse words in Spanish and French using Mozilla Common
Voice. \autoref{tab:cross_lingual_swears} demonstrates that \li{3584}
is similarly capable of steering in all three languages, here shown avoiding swear words in the resultant transcript from the steered latent vector. This cross-lingual generalization is particularly notable for French, which was entirely absent from the SAE training data.

\begin{table*}[t]
\centering
\footnotesize
\caption{French and Spanish samples negatively steered using \li{3584} to remove profanity.}
\label{tab:cross_lingual_swears}
\begin{tabular}{|p{0.45\textwidth}|p{0.45\textwidth}|}
\hline
    \textbf{Original} & \textbf{Steered} \\
\hline
\hline
\multicolumn{2}{|c|}{\textit{French}} \\
\hline
    littéralement Darwin est une \textbf{salope} & L'étérale-midi L'étérale-midi \\
\hline
    Ou ça ? Dans ton \textbf{cul} ? & ou ça, dans tant que du tout \\
\hline
    C'est vraiment un gros \textbf{con}, on l'y en a & C'est vraiment un grand congrès \\
\hline
    Mais ce qui va \textbf{chier}, dehors & Mets ce qui va chez deux ors \\
\hline
    Tu vas voir \textbf{ta gueule} ! & Tu vas voir ta gagne \\
\hline
\hline
\multicolumn{2}{|c|}{\textit{Spanish}} \\
\hline
    Lo he oído, \textbf{pendejo} & Lo he oído en dejo \\
\hline
    Con una mano me basta para patiarte el \textbf{culo} & Con una mano me basta para apatirte el color \\
\hline
    Entras en este ciclo de \textbf{zorra} & entre en este ciclo de sorra \\
\hline
    ¡Cómo mucho! ¡Dios, \textbf{Joder}! & Como mucho. Díos, Jadera \\
\hline
    más seguro es en la cabina, será \textbf{idiota} & Más seguro en la cabina, se ha ido a la idea \\
\hline
\end{tabular}
\end{table*}

\subsection{Steering Results Overview}

 The results of \autoref{tab:steerability} demonstrate two main inconsistencies with steering. The first is that lower-level latents, such as phonetic information, seem to be more difficult to steer than higher-level latents, such as semantic information. Further, steering a feature on with a positive value seems to be easier than steering a feature off with a negative value. We return to this point in the discussion section.

\begin{table}[htbp]
\centering
\footnotesize
\caption{Summary of causal steering outcomes by level of the linguistic hierarchy.}
\label{tab:steerability}
\begin{tabular}{|l|l|c|c|}
\hline
\textbf{Level} & \textbf{Latent} & \textbf{Negative} & \textbf{Positive} \\
\hline
Phonetic & 6373 & No & No \\
Lexical & 1445 & No & Yes \\
Syntactic & 1163 & No & Yes \\
Morphological & 29 & No & Yes \\
Semantic (Numbers) & 1492 & No & Yes \\
Semantic (Profanity) & 3584, 104 & Yes & Yes \\
\hline
\end{tabular}
\end{table}

\section{Discussion and Conclusion} \label{sec:discussions}

The present study demonstrates that Whisper encodes a variety of linguistic information ranging from low-level acoustic information to higher level semantic information. Specifically, we find that the Whisper encoder represents information including phonetic, lexical, syntactic and morphological, and semantic information.

Why might the encoder represent information that exceeds what transcription strictly requires? One possibility is that since phonetically similar utterances can correspond to very different intended meanings, resolving this kind of ambiguity may require drawing on information beyond the acoustic signal alone. Critically, different levels of the hierarchy documented here could each help resolve a different source of local ambiguity. Lexical information can help distinguish between specific words that are acoustically similar but orthographically and semantically distinct, such as \textit{flower} and \textit{flour}. Morphological information can help resolve ambiguity at the level of word-internal structure, for instance whether a given sound sequence marks an inflectional ending or is simply part of the stem, as in the homophonous \textit{passed} (\textit{pass} plus the past-tense suffix \textit{-ed}) and \textit{past} (a single, non-inflected morpheme). Semantic information, in turn, can help resolve ambiguity at the level of the intended meaning of an utterance as a whole, such as determining whether \textit{bank} refers to a financial institution or the edge of a river, in cases where multiple interpretations remain plausible even once lexical and morphological information have been taken into account. Under this view, the rich hierarchy of representations documented here is not incidental: each level may serve as a distinct source of disambiguating evidence, with different levels resolving different kinds of local uncertainty in the acoustic signal.

Another possibility is that encoding redundant signals may help reduce mistranscriptions. Indeed, linguists have posited that the integration of multiple information sources allows speech perception to be more robust to errors \citep{gibson2013rational}. Noisy-channel processing models, for example, propose that listeners overcome ambiguous or degraded signals by actively combining multiple cues, phonetic, lexical, semantic, and contextual, to recover the most likely intended message \citep{gibson2013rational, houghton2024noisychannel, houghton2025bolts}. Whisper, having been trained on naturalistic and noisy audio at scale, may have converged on a similar strategy, encoding seemingly redundant information across multiple levels of abstraction as a coping mechanism against exactly this kind of ambiguity.

Beyond explaining why the encoder might represent this information, these findings also speak to broader debates within linguistics. One such debate concerns the extent to which language models rely on memorized patterns as opposed to genuine abstraction \citep{houghton2025multi, houghton2025role}. A model that merely memorizes could, in principle, represent isolated instances of profane words or number words without ever forming a category that unifies them. The existence of a single latent that reliably activates across many distinct profane words, and a separate latent that activates across many distinct numbers, is itself evidence that Whisper has formed abstract, category-level representations rather than a collection of item-specific memorized patterns.

The cross-lingual results strengthen this conclusion. The profanity latent's activation for French is particularly informative because French was absent from the SAE's training data: the SAE could not have learned a French-specific profanity pattern. If the latent's behavior were instead an artifact of surface-level correlations specific to English, for instance, phonetic or orthographic patterns common to English swear words, it should not generalize to a language with entirely different phonetic and orthographic patterns. Its generalization to French, therefore, suggests that the underlying category is represented at a level of abstraction within Whisper's encoder that is not tied to any particular language's surface form, rather than being an artifact of English-specific patterns in the training data.

Finally, the steering results suggest that low-level features may be more difficult to steer than higher-level features. This is possibly due to the encoder's function: while the encoder does contain high-level semantic information, it may encode lower-level features more redundantly (such as representing a triphone in terms of its triphone, diphones, and monophones) allowing the encoder to be robust to noise that interferes with part of the signal. Indeed, we find many different latents corresponding to varying degrees of granularity (see \appref{appx:latent_samples}). This may also explain why steering a feature off is more difficult than steering a feature on. When a feature is steered off, there are often still other signals as to the identity of that feature. The decoder may still be able to use this information to recover from the degradation of the signal. This is further evidenced by the pattern of substitutions observed by steering: when semantic information is turned off (negative steering), the decoder falls back to a phonetically similar word with no particular semantic constraint (e.g.,\ \textit{fuck}$\rightarrow$\textit{Falk}), whereas when it is turned on instead (positive steering), the decoder selects a phonetically similar word that also conforms, at least loosely, to the target semantic category (e.g.,\ \textit{federal courts}$\rightarrow$\textit{50 crores}, itself a number). This suggests ASR output is more tightly constrained by the encoder than LLM generation is: regardless of steering direction, phonetic information persists and continues to guide the decoder's transcription. That being said, this argument is preliminary. We note that this account is inferred from the pattern of results rather than independently verified. We leave further examination of this to future work.

Altogether, the results of the present study demonstrate the richness of the representations embedded in the Whisper encoder, spanning low-level acoustic and phonetic information through lexical, morphological, and syntactic information, up to semantic information.

\FloatBarrier
\bibliographystyle{unsrtnat}
\bibliography{main}

\appendices

\onecolumn

\newpage
\section{Dataset Composition and Language Identity Latent}
\label{appx:extra_tables}

\subsection{Training Set Composition}
\autoref{tab:trainset} gives the exact file count contributed by each source to the SAE training set.

\begin{table}[htbp]
\centering
\footnotesize
\caption{Training set composition by source, 646,769 files total. ``Internal'' rows
are audio not publicly available; the largest of these (the synthetic
TTS clone) uses a single synthesized voice reading the LJSpeech transcripts, not
real human recordings.}
\label{tab:trainset}
\begin{tabular}{lrr}
\toprule
\textbf{Source} & \textbf{N} & \textbf{\%} \\
\midrule
LibriSpeech & 519,746 & 80.36\% \\
Spanish accent corpora (SLR61, SLR71--75) & 19,542 & 3.02\% \\
Mozilla Common Voice (English) & 19,051 & 2.95\% \\
Zamia Speech & 12,068 & 1.87\% \\
LJSpeech & 10,170 & 1.57\% \\
Internal: coherence-task recordings & 11,863 & 1.83\% \\
TEDx Spanish Corpus (SLR67) & 9,007 & 1.39\% \\
West Point Heroico Spanish (SLR39) & 8,974 & 1.39\% \\
VoxCeleb 1 & 8,726 & 1.35\% \\
Internal: synthetic TTS clone of LJSpeech text & 14,109 & 2.18\% \\
Internal: resampled/other & 4,483 & 0.69\% \\
Internal: 8kHz telephony-style audio & 3,438 & 0.53\% \\
Internal: spoken-digit recordings & 1,852 & 0.29\% \\
Internal: other (call-routing, survey, misc.) & 2,947 & 0.46\% \\
Musan & 793 & 0.12\% \\
\bottomrule
\end{tabular}
\end{table}

\subsection{Analysis Set Composition}
\autoref{tab:analysisset} summarizes the composition of the dedicated analysis set described in \autoref{sec:analysis_set}.

\begin{table}[htbp]
\centering
\footnotesize
\caption{Analysis set composition.}
\label{tab:analysisset}
\begin{tabular}{llr}
\toprule
\textbf{Dataset} & \textbf{Source} & \textbf{N} \\
\midrule
LJSpeech & LJSpeech-1.1 & 13,084 \\
Common Voice (English) & CV Corpus 13.0 & 7,830 \\
Common Voice (English, profanity) & CV Corpus 13.0 & 331 \\
Common Voice (Spanish) & CV Corpus 13.0 & 7,500 \\
\midrule
\textbf{Total} & & \textbf{28,414} \\
\bottomrule
\end{tabular}
\end{table}

\subsection{Language Identity Latent}
The effect of language was examined by comparing latent activations across Spanish and English samples in the analysis set. A simple analysis evaluated whether a single activation of any latent, in any frame of an utterance, could discriminate between English and non-English utterances. This approach identified \li{5106}, which classifies non-English samples with a precision of $77.6\%$ and a recall of $91.1\%$. Causal validation via positive steering is discussed in \autoref{sec:discussions}; briefly, positively steering \li{5106} did not flip Whisper's own language classification at any tested magnitude. Negative steering was not tested, since the analysis set used for steering contains no genuine non-English content for \li{5106} to be suppressed on.

\newpage
\section{Training Configuration}
\label{appx:training_config}

The SAE was trained with the Adam optimizer \citep{kingma2014adam} (learning rate $10^{-4}$,
$\beta_1=0.9$, $\beta_2=0.999$, no weight decay), a batch size of 16 utterances, and no
L1 sparsity penalty ($\lambda=0$), relying on the TopK activation alone to enforce
sparsity. Training used early stopping on a held-out validation split, monitored every
one-fifth of an epoch, with a patience of 15 validation checks (3 epochs) and a minimum
improvement threshold of $10^{-4}$ on the validation reconstruction loss. Under this
criterion, training converged after 9 epochs (400,000 optimizer steps) over the training set. Training was performed on a single NVIDIA A100 80GB PCIe GPU.

\newpage
\section{Ground-Truth Labeling Procedures}
\label{appx:labeling}

To evaluate precision and recall for each hypothesized feature (\autoref{sec:feature_id}), ground-truth labels indicating whether a given feature was present in each file were required. The annotation procedure differed by feature type.

\subsection{Phonetic Features}
Ground-truth phone-level labels were obtained from existing TextGrid annotations for a subset of the LJSpeech corpus, which align audio to human-annotated phone-level transcriptions (\autoref{sec:analysis_set}).

\subsection{Morphological Features}
Candidate instances of the hypothesized morphological structure (for example, words containing a given suffix) were first identified using an automated morphological tagger. Each candidate was then labeled by a large language model, and the resulting labels were verified by a human annotator.

\subsection{Lexical Features}
Ground-truth labels for the presence of a specific word were obtained through lemma-tagging, identifying occurrences of the target lemma in each file's transcript.

\subsection{Semantic Features}
Ground-truth labels for the semantic categories examined in this work (numbers and profanity) were obtained through human curation: annotators selected files in which words belonging to the target category occurred.

This curation procedure is a plausible source of the comparatively lower precision observed for the semantic features (\autoref{sec:semantic}). A word belonging to the target category is not always used in a manner consistent with the hypothesized feature: a curated word can appear in a context where it is not functioning as the intended category (for example, a word normally associated with profanity used non-profanely, or a word associated with numbers without itself denoting a number), which would cause a genuinely correlated latent to be scored as a false positive. Conversely, a latent may fail to activate for some instances of the target category due to feature splitting, a phenomenon independently documented elsewhere in this paper (\autoref{sec:syntactic_and_morpho}), whereby closely related instances of a feature are represented by distinct, more narrowly tuned latents, depressing recall. The curation procedure used for the semantic features is coarser than the annotation procedures used for phonetic, morphological, and lexical features, which may explain why the semantic latents show comparatively weaker precision despite plausibly representing a genuine underlying category, a possibility further supported by the causal steering results (\autoref{sec:profanity}) and the cross-lingual generalization reported there.

\newpage
\section{Domain Space}
\label{appx:data_trans}

Let $E: \mathcal{X} \rightarrow \mathcal{Z}$ denote a fixed Whisper encoder, mapping inputs from a
high-dimensional space $\mathcal{X}$ to a latent space $\mathcal{Z}$. The encoder's weights
define an image manifold $\mathcal{M} = \{E(x) : x \in \mathcal{X}\} \subset \mathcal{Z}$, a
structured subset of $\mathcal{Z}$ whose geometry is entirely determined by $E$. A sparse
autoencoder trained on the outputs of $E$ operates exclusively in $\mathcal{Z}$-space
and therefore has access only to sampled points on $\mathcal{M}$. Critically, $\mathcal{M}$
is invariant to any choice of SAE training data: an ablation that removes a semantic subset
of $\mathcal{X}$ modifies the sampling measure over $\mathcal{M}$ but cannot alter
$\mathcal{M}$ itself. The SAE cannot recover structure that $E$ does not encode, nor can it
fail to recover structure that $E$ does encode due to ablation alone, provided coverage of
the manifold is sufficient.

An analogous invariance holds for SAE hyperparameters. The latent space dimensionality and
sparsity constraint parameterize a family of approximations
$\varphi_\theta(\mathcal{M})$ to the manifold, but varying $\theta$ does not alter
$\mathcal{M}$. Increasing latent dimensionality improves the granularity of the
decomposition; relaxing sparsity allows greater feature entanglement per activation. Both
affect approximation fidelity, not the underlying representational geometry. Consequently,
ablation studies over training data composition and hyperparameter searches are properly
understood as studies of approximation error under a fixed target $\mathcal{M}$, rather than
as explorations of what structure is in principle recoverable from $E$. The object of
interest, the geometry of $\mathcal{M}$ as determined by the encoder (representing the discovered features), is not a
variable in either type of experiment.

This argument applies regardless of the exact degree of overlap between the SAE's training
data and the analysis set used to evaluate it (\autoref{sec:analysis_set}): whether a
specific analysis file was or was not among the SAE's training examples has no bearing on
$\mathcal{M}$, since $\mathcal{M}$ is a property of the fixed, frozen encoder $E$ alone, not
of the SAE or its training data. What follows from this argument is narrower than it may
first appear: it establishes that structure recoverable by the SAE is bounded by what $E$
already encodes, regardless of training/analysis overlap, but it does not by itself rule out
a distinct concern, that the SAE could fit its decomposition more tightly to files it happened
to train on, making labeled latents look more precise on those files than they would on
genuinely novel input. That concern is addressed empirically rather than argued away here: as
noted in \autoref{sec:analysis_set}, a majority of the analysis set, and effectively all of
the Common Voice portion used for the cross-lingual results in \autoref{sec:profanity}, was
never seen during SAE training, and the reported results hold on that held-out majority.

\newpage
\section{Automated Labeling Procedure}
\label{appx:auto_label}

To supplement the manual feature analysis, all latents were labeled using GPT-OSS 120B. For each latent, the 100 files closest to the median activation strength were selected as representative examples, and up to 20 were presented to the model with activated character spans marked by asterisks (e.g., \texttt{grad*ual*ly}). The model was prompted to assign each example one of seven categories: phonetic, orthographic, morphological, lexical, semantic, syntactic, or diffuse. The model subsequently returned a structured JSON response containing a label, explanation, confidence level, and category. Only high-confidence labels were retained; lower-confidence responses were assigned the \textit{diffuse} label with the leading hypothesis preserved. A random subset of 100 of these latents was manually evaluated to determine the quality of the automated labeling. The automatically labeled latents yielded an accuracy of 76\%.

\newpage
\section{Latent Samples}
\label{appx:latent_samples}
{\fontsize{8}{11}\selectfont
\begin{longtable}{rlr}
\toprule
latent\_id & label & correct \\
\midrule
\endfirsthead
\toprule
latent\_id & label & correct \\
\midrule
\endhead
\midrule
\multicolumn{3}{r}{Continued on next page} \\
\midrule
\endfoot
\bottomrule
\endlastfoot
147  & sentence-initial word (first word of each sentence) & 1 \\
570  & voiceless alveolar fricative /\textipa{s}/ (spelled s, c, x, ss, etc.) & 1 \\
604  & detects unstressed vowel (schwa-like /\textipa{@}/) & 1 \\
668  & presence of the /\textipa{3:}/ (schwa-r) vowel sound (as in ``er/ir/ur'' syllables) & 0 \\
722  & words containing the digraph ``sh'' or ``ch'' & 0 \\
730  & words containing the /\textipa{dZ}/ (j/soft-g) sound & 1 \\
994  & words containing the /\textipa{k}/ (hard c/k/ck) sound & 1 \\
1052 & back-rounded vowel /\textipa{O}/ (open-mid back rounded vowel) & 1 \\
1288 & words with the long-A /\textipa{eI}/ vowel (face/race rhyme family) & 0 \\
1579 & detects the word ``hundred'' (including plural and compound forms) & 1 \\
1626 & presence of a nasal consonant (/\textipa{n}/ or /\textipa{m}/) & 1 \\
1799 & detects the orthographic bigram ``so'' & 1 \\
1822 & presence of final /\textipa{s}/ sound & 0 \\
1992 & detects the Latin root ``clud/clus'' (include, exclude, conclude, etc.) & 1 \\
2127 & voiceless velar stop /\textipa{k}/ phoneme & 1 \\
2151 & determiner (articles, demonstratives, numerals) & 1 \\
2219 & detects words containing the letter ``f'' & 1 \\
2318 & detects the lateral consonant /\textipa{l}/ (letter l) & 1 \\
2347 & detects the /\textipa{s}/ fricative (s-sound) & 1 \\
2353 & presence of the long-A diphthong /\textipa{eI}/ (as in ``state'', ``great'', ``candidate'', ``tolerate'') & 0 \\
2561 & presence of the velar nasal /\textipa{N}/ (ng/nk sound) & 1 \\
2875 & initial stop-liquid consonant clusters (e.g., /\textipa{pr}/, /\textipa{br}/, /\textipa{fr}/, /\textipa{dr}/) & 0 \\
3202 & words containing the /\textipa{s}/ sound (voiceless alveolar fricative) & 1 \\
3559 & words ending in ``s'' (plurals/possessives) & 1 \\
3652 & dental fricative /\textipa{T}/ (th) sound & 0 \\
3714 & detects \textendash ed past-tense/participle suffix & 1 \\
3733 & words containing a consonant cluster (two adjacent consonants) & 0 \\
3920 & activation on short function words and morphemes (auxiliaries, prepositions, articles) & 0 \\
4413 & activation on words containing the alveolar stop /\textipa{t}/ (or /\textipa{d}/) & 0 \\
4555 & detects the /\textipa{n}/ nasal sound & 1 \\
4808 & first vowel (early vowel sound) of a word & 0 \\
4948 & short-i vowel /\textipa{I}/ (high front lax vowel) & 0 \\
5021 & detects back-rounded vowel /\textipa{o}/ (e.g., `o'/`ow' sound) & 1 \\
5045 & presence of the /\textipa{t}/ phoneme (letter t) & 1 \\
5060 & detects the morpheme ``with-'' (with, within, without, forthwith) & 1 \\
5134 & presence of a sibilant fricative (/\textipa{s}/, /\textipa{z}/, /\textipa{S}/) & 1 \\
5237 & detects the prefix ``after-'' (e.g., after, afterwards, thereafter, afterwork, afternoons) & 1 \\
5268 & detects rhotic /\textipa{r}/ sound (letter r) & 1 \\
5349 & presence of the /\textipa{s}/ sound (voiceless alveolar fricative) & 1 \\
5352 & presence of /\textipa{s}/ or /\textipa{z}/ (alveolar fricative) sound & 1 \\
5629 & detects the /\textipa{st}/ consonant cluster & 1 \\
5633 & detects the /\textipa{v}/ (voiced labiodental fricative) sound & 1 \\
6079 & detects common Latin-derived affixes (com-, de-, re-, dis-, su-, -es) & 0 \\
6136 & detects the -ing participial/gerund suffix & 1 \\
6206 & first word after punctuation (clause-initial) & 0 \\
6334 & detects the /\textipa{tS}/ (ch) sound & 1 \\
6411 & negation particles (not / no) & 1 \\
6413 & presence of bilabial nasal /\textipa{m}/ sound & 1 \\
6417 & presence of the schwa vowel /\textipa{@}/ in the word & 0 \\
6692 & function words -- articles, prepositions, demonstratives, pronouns & 1 \\
6775 & rhotic consonant /\textipa{r}/ (\textipa{r}) activation & 1 \\
7289 & detects the /\textipa{r}/ consonant & 1 \\
7507 & detects the lateral /\textipa{l}/ sound & 1 \\
7781 & detects the /\textipa{v}/ (voiced labiodental fricative) sound & 1 \\
7867 & short function words (e.g., ``of'', ``a'', ``by'', ``off'') & 1 \\
7954 & detects the morpheme ``hundred'' (numeric multiplier) & 1 \\
8229 & presence of a nasal consonant (e.g., /\textipa{n}/ or /\textipa{m}/) in the word & 1 \\
8310 & detects the /\textipa{ru:}/ (long-u) sound & 1 \\
8365 & presence of the lateral /\textipa{l}/ sound (letter L) & 1 \\
8433 & words containing a voiced labial consonant (/\textipa{b}/ or /\textipa{v}/) & 1 \\
8555 & third-person plural pronoun ``them'' and its inflections & 1 \\
8748 & definite article ``the'' (including capitalised forms) & 1 \\
8765 & words containing the /\textipa{saI}/ (s-ai) sound & 1 \\
8782 & closed-class function words (prepositions, conjunctions, determiners) & 0 \\
8785 & affricate consonant /\textipa{tS}/ or /\textipa{dZ}/ (ch/j) & 1 \\
9293 & detects the /\textipa{nVm}/ (or /\textipa{nom}/) sound cluster & 1 \\
9722 & presence of the high front vowel /\textipa{i}/ (ee/eye sound) & 1 \\
9779 & detects the \textendash ly suffix (adverbial/derived \textendash ly words) & 1 \\
9794 & presence of the /\textipa{s}/ (alveolar fricative) phoneme & 1 \\
10429 & detects presence of the /\textipa{n}/ nasal sound (letter n) & 1 \\
10523 & definite article ``the'' (and similar determiners) & 1 \\
10668 & punctuation (comma, semicolon, dash) and following word & 0 \\
10724 & detects the word ``time'' (including plural, capitalized, and compounds) & 1 \\
11705 & detects the rhotic /\textipa{r}/ sound & 1 \\
12424 & words containing the diphthong /\textipa{aI}/ (mind, mine, might, line, turbine, etc.) & 1 \\
12477 & detects rhotic /\textipa{r}/ sound (letter r) & 0 \\
12689 & presence of the /\textipa{kl}/ consonant cluster & 1 \\
12697 & sentence-initial word (first token) & 1 \\
12844 & presence of the letter ``o'' in a word & 0 \\
12963 & words containing the /\textipa{On}/-/\textipa{wVn}/ sound (e.g., one, won, on, wan, want, worn) & 1 \\
13064 & detects the /\textipa{In}/ (in) sound or grapheme & 1 \\
13432 & words ending in a nasal consonant (/\textipa{m}/, /\textipa{n}/, /\textipa{N}/) & 1 \\
13696 & words containing the postalveolar affricate /\textipa{tS}/ or /\textipa{dZ}/ sound & 1 \\
13724 & initial consonant clusters (st, tr, cr, pr, sh, cl, gr, etc.) & 0 \\
14001 & words with the /\textipa{eIk}/ sound (make, take and inflections) & 1 \\
14039 & detects the word ``and'' (coordinating conjunction) & 1 \\
14157 & closed-class function words (prepositions, articles, auxiliaries, conjunctions) & 0 \\
14373 & detects the /\textipa{wel}/ phoneme cluster (e.g., ``well'', ``twelve'', ``Welsh'', etc.) & 1 \\
14601 & words containing the alveolar fricative /\textipa{s}/ (or its voiced counterpart /\textipa{z}/) & 1 \\
14752 & words containing a voiced sibilant (/\textipa{z}/ or /\textipa{Z}/) & 1 \\
14770 & detects bilabial stop consonants (/\textipa{b}/ or /\textipa{p}/) & 1 \\
14851 & initial vowel-onset words & 0 \\
14869 & first-person possessive determiner ``my'' (including ``myself'') & 1 \\
15083 & rhotic /\textipa{O:r}/ vowel (or/ore/oor) sound & 1 \\
15123 & voiced labiodental fricative /\textipa{v}/ & 1 \\
15387 & presence of a labial consonant (b / p / v) & 1 \\
15606 & onset of primary stress syllable (initial consonant or cluster) & 0 \\
15749 & short function words (prepositions, conjunctions, particles) & 1 \\
15915 & presence of the /\textipa{S}/ (sh) sound & 1 \\
15939 & presence of alveolar liquid consonants (/\textipa{l}/ or /\textipa{r}/) & 0 \\
\end{longtable}
}

\newpage
\section{Non-Linguistic Latents}
\label{appx:others}
\subsection{Noise}
As previously mentioned, noise information has been shown to be present in the
Whisper encoded data, however it was not clear how this information would be
encoded within the SAE latent space as an activation. Noisy samples were sourced
from ESC-50 which contains $50$ distinct types of non-speech noise
\citep{piczak2015dataset}. By comparing voiced and non-voiced samples, the
activation of \li{15019} clearly discriminated the two classes. A predictor was
created according to the following logic:

\[
pred=
\begin{cases}
\text{if } \ell_{15019} > 0 & \text{noise} \\[2pt]
\text{else}                 & \text{voice}
\end{cases}
\]

\begin{figure}[htbp] \centering
        \includegraphics[width=0.5\linewidth]{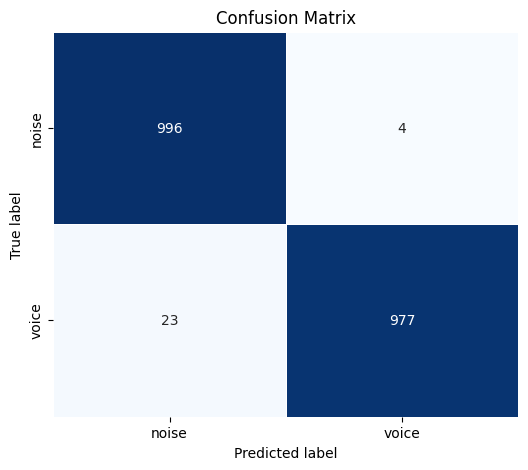}
        \caption{\Li{15019} is associated with noise only audio, the latent index can
            effectively separate noise-only audio from speech audio.}
        \label{fig:noise}

\end{figure}

\autoref{fig:noise} shows the confusion matrix created using this predictor to
compare to the ground truth. This predictor has a precision of $98.1\%$ and a
recall of $99.8\%$ for the noise audio samples, indicating that this latent
fires on essentially all noise samples while rarely firing on clean speech.

\subsection{Positional Features}
\label{sec:positional}
\begin{figure}[htbp] \centering
    \includegraphics[width=0.55\columnwidth]{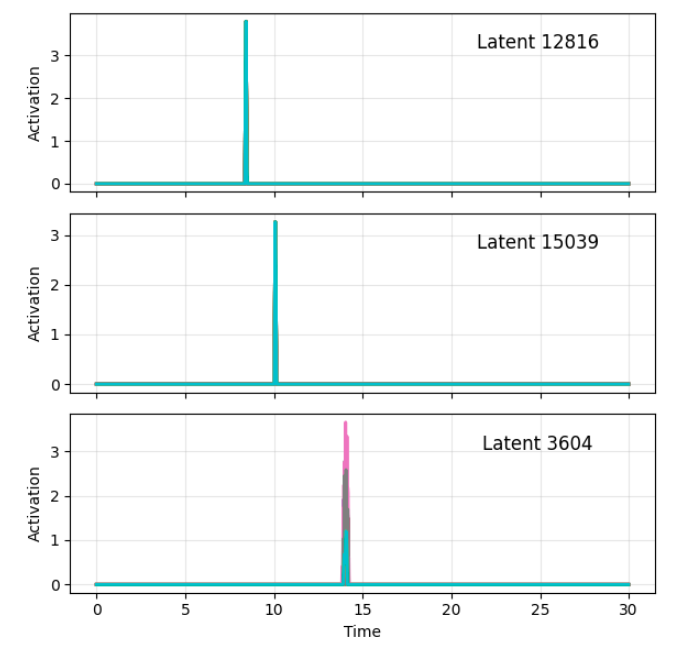}
    \caption{Three different positional latents showing activations on 20
    different audio files. These positional latents are very temporally
constrained and consistent between distinct files.}
\label{fig:positional}
\end{figure}

The Whisper encoder uses a positional encoding which is added to the input features,
therefore some positional information likely exists in the encoding. However, since
the SAE encodes all information as discrete latent indices, it is not clear how
the positional information would manifest. \autoref{fig:positional} shows
three examples of positional latent indices, all of which are tightly located
temporally. \Li{12816} has a mean activation timestamp of 8.44 seconds and a
standard deviation of 0.19. \Li{15039} is at 10.09 seconds on average with an SD
of 0.04, and \li{3604} activates at 14.04 seconds with an SD of 0.37. This positional
representation is potentially useful for the decoder to determine word order since the encoding
is passed to the decoder with cross-attention.

\end{document}